%% file: 0_emotion_language_arxiv.tex
\pdfoutput=1
\documentclass[sigconf, nonacm]{acmart}

\AtBeginDocument{%
  \providecommand\BibTeX{{%
    \normalfont B\kern-0.5em{\scshape i\kern-0.25em b}\kern-0.8em\TeX}}}

\setcopyright{none}
\renewcommand\footnotetextcopyrightpermission[1]{}

\usepackage{graphicx}
\usepackage{booktabs}
\usepackage{amsmath}

\begin{document}

\title{LG-GER: Language-Guided Group Emotion Recognition via Multimodal Evidence Distillation}

\author{Ahmed Shehab Khan}
\affiliation{%
  \institution{University of South Carolina}
  \city{Columbia}
  \state{South Carolina}
  \country{USA}}
\email{akhan@email.sc.edu}

\author{Zhiyuan Li}
\affiliation{%
  \institution{University of South Carolina}
  \city{Columbia}
  \state{South Carolina}
  \country{USA}}
\email{zhiyuanl@email.sc.edu}

\author{Yan Tong}
\affiliation{%
  \institution{University of South Carolina}
  \city{Columbia}
  \state{South Carolina}
  \country{USA}}
\email{tongy@cse.sc.edu}

\renewcommand{\shortauthors}{Khan, Li, and Tong}

\begin{abstract}
  Inferring the collective emotional state of a group of people from a single image, a task known as group emotion recognition (GER), requires integrating spatially distributed cues such as faces, poses, interactions, and scene context. Current methods rely on detector-driven multi-stream pipelines. These are trained with only image-level supervision that lacks guidance on which regions matter or how strongly each contributes. We propose LG-GER, a language-guided distillation framework that uses a multimodal large language model (MLLM) to generate dense, spatially grounded evidence, i.e., bounding boxes paired with emotion signals and confidence scores, for the training images. This structured evidence is distilled into a single vision-language model (VLM) backbone through four complementary losses: classification, region-text grounding, spatial emotion, and spatial confidence regression. At inference, LG-GER requires no detectors, no MLLM, and no multi-stream fusion, making GER practical for real-time and resource-constrained deployment. LG-GER has been evaluated on two benchmark GER datasets (GroupEmoW and GAF~3.0) and achieves competitive or superior results compared to state-of-the-art methods that require detection and multi-stream processing at inference.
\end{abstract}


\maketitle

\input{1_introduction}
\input{2_related_work}

\input{3_methodology}
\input{4_experiment}
\input{5_conclusion}

\clearpage
\bibliographystyle{ACM-Reference-Format}
\bibliography{references-strings,references}

\end{document}

%% file: 1_introduction.tex
\section{Introduction}
\label{sec:intro}

\begin{figure}[t]
  \centering
  \includegraphics[width=0.88\columnwidth]{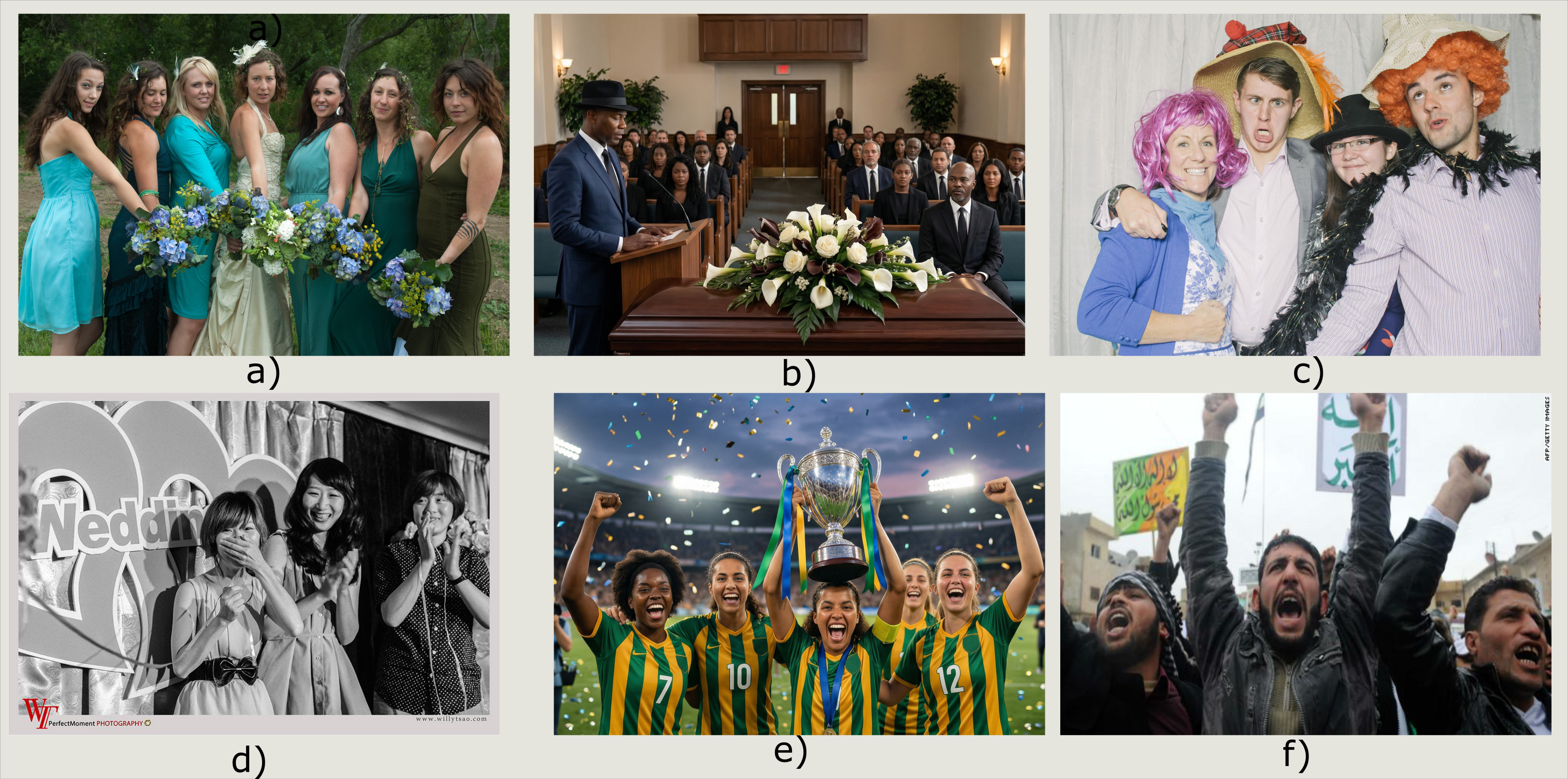}
  \caption{\small{The ambiguity of isolated visual cues in group emotion recognition. (a,\,b)~Flowers appear at both a wedding (positive) and a funeral (negative). (c,\,d)~Facial expressions alone are insufficient: costumes obscure intent in~(c), while scene context reveals positive emotion despite an occluded face in~(d). (e,\,f)~Raised fists signal celebration with a trophy~(e) but anger during a protest~(f). Reliable GER requires integrating faces, objects, scene layout, and interactions.}}
  \label{fig:ger_complexity}
\end{figure}

Understanding the collective emotional state of a group from a single image requires more than reading individual faces, as they alone do not provide sufficient information. Facial expressions, body language, held objects, interpersonal interactions, and the surrounding scene all contribute evidence, but relying on any single cue alone can be misleading. As illustrated in Figure~\ref{fig:ger_complexity}, the same visual element can signal opposite emotions depending on context. Inferring a group-level emotion label by integrating such heterogeneous and potentially contradictory cues is known as \emph{Group Emotion Recognition} (GER). Humans resolve these ambiguities by jointly weighing cue type, context, and inter-person dynamics~\cite{barrett2011context}, but computational models have struggled to replicate this compositional reasoning.

The dominant computational paradigm decomposes GER into multi-stream pipelines: dedicated detectors extract faces, bodies, or objects; separate backbones encode each stream; and a late-stage module fuses the resulting features. Despite steady benchmark progress, this paradigm suffers from three persistent limitations. First, detector failures propagate into classification errors. Second, maintaining multiple detection and encoding networks increases engineering overhead, adds potential failure points, and incurs additional inference cost. Third, a single image-level label lacks region-level guidance.

\begin{figure*}[t]
  \centering
  \includegraphics[width=\textwidth]{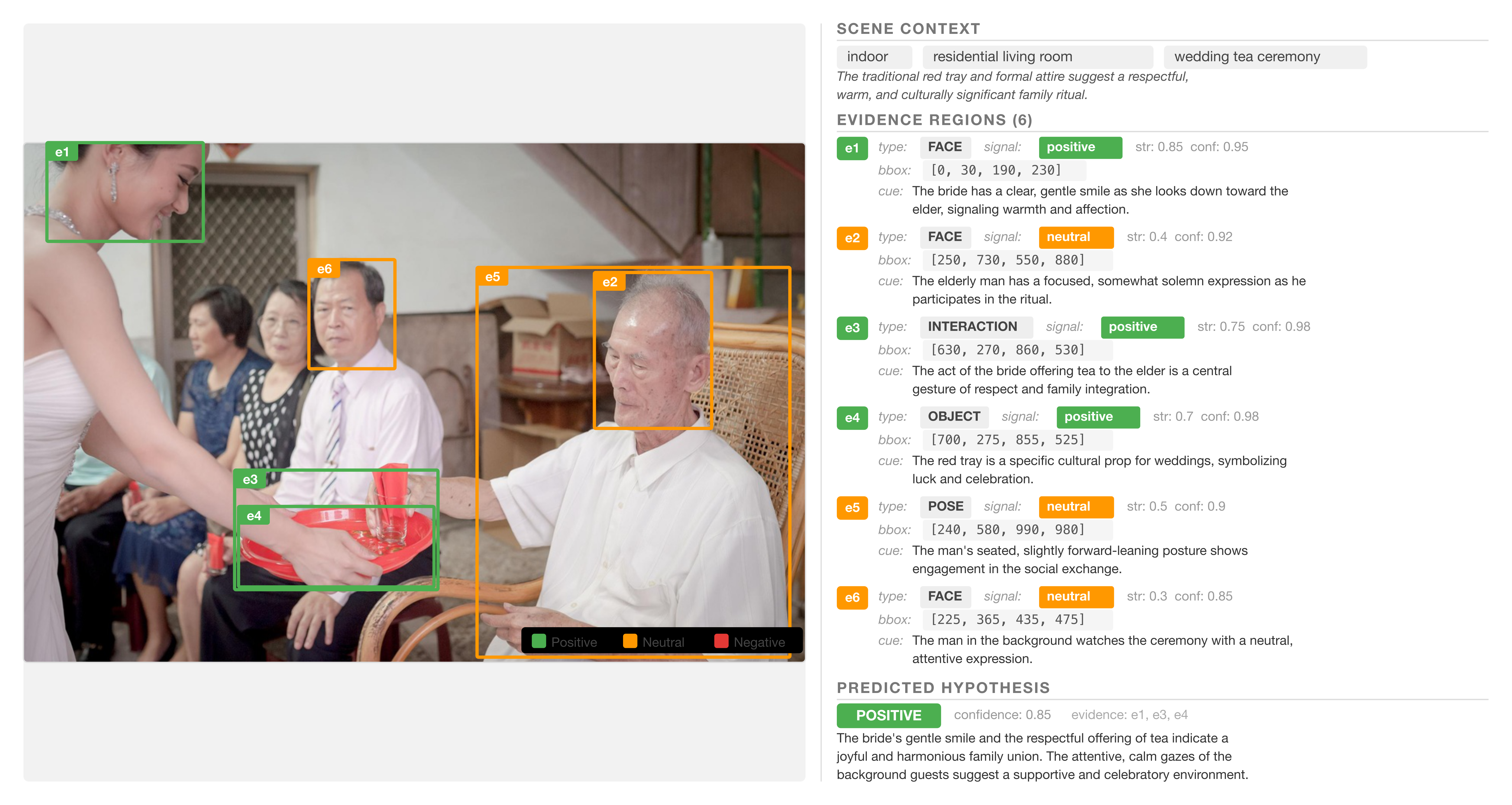}
  \caption{\small{An example of MLLM-generated annotations. Each evidence region includes a bounding box, type, cue description $c_i$, emotion signal $s_i$, and signal strength $\tau_i$.}}
  \label{fig:annotation_example}
\end{figure*}

Multimodal large language models (MLLMs) offer a potential remedy. Models such as Gemini 3.0 Flash~\cite{gemini2025flash} can identify faces, gestures, objects, and scene context in images, articulating the kind of multi-cue reasoning that GER demands. However, deploying billion-parameter models at inference is impractical, and fine-tuning them for a downstream task like GER is computationally expensive. Our key insight is that an MLLM can instead serve as an \emph{offline annotator}, generating structured spatial evidence, i.e., bounding boxes, emotion signals, and confidence scores, for the training images (Figure~\ref{fig:annotation_example}). This evidence is then \emph{distilled} into a lightweight VLM backbone, which learns to outperform the MLLM's own zero-shot classification through spatially grounded supervision.

We introduce \textbf{LG-GER} (Language-Guided Group Emotion Recognition), a framework that transfers structured MLLM evidence into a single VLM backbone via four complementary losses. At inference, LG-GER requires no detection, no MLLM, and no multi-stream fusion. Our contributions are:
\begin{itemize}
    \item A \textbf{scalable MLLM evidence pipeline} that generates dense, spatially grounded, region-level supervision for GER training images; and
    \item A \textbf{multi-signal distillation framework} with four complementary losses and an \textbf{Emotion Adapter} for emotion-discriminative region-text grounding. 
\end{itemize}

The LG-GER has been evaluated on two GER benchmark datasets, showing that this distillation approach achieves competitive or superior results compared to state-of-the-art multi-stream methods.

%% file: 2_related_work.tex
\section{Related Work}
\label{sec:related_work}

\subsection{Individual Emotion Recognition}
\label{sec:rw_ier}

Individual emotion recognition (IER) has progressed from controlled laboratory settings with posed expressions to large-scale in-the-wild benchmarks, driven by increasingly powerful CNN and transformer architectures~\cite{wang2024fersurvey}. Beyond facial expressions, body pose and scene context provide complementary cues for emotion prediction~\cite{kosti2017emotion}, and scene context alone can sometimes outperform the face itself~\cite{chen2016emotion}. More recently, MLLMs have been applied to individual emotion tasks through instruction tuning~\cite{li2024emola}, multi-modal fusion~\cite{cheng2024emotionllama}, and chain-of-thought reasoning~\cite{lan2025expllm}, establishing a growing intersection between foundation models and affective computing~\cite{shou2025mllmemotion}.

These works establish two principles that motivate our approach: emotion perception benefits from integrating multiple cues beyond the face, and surrounding context can be more informative than facial expression alone. However, all IER methods remain confined to single individuals and cannot address the compositional, multi-person nature of group emotion recognition.

\begin{figure*}[t]
  \centering
  \includegraphics[width=0.9\textwidth]{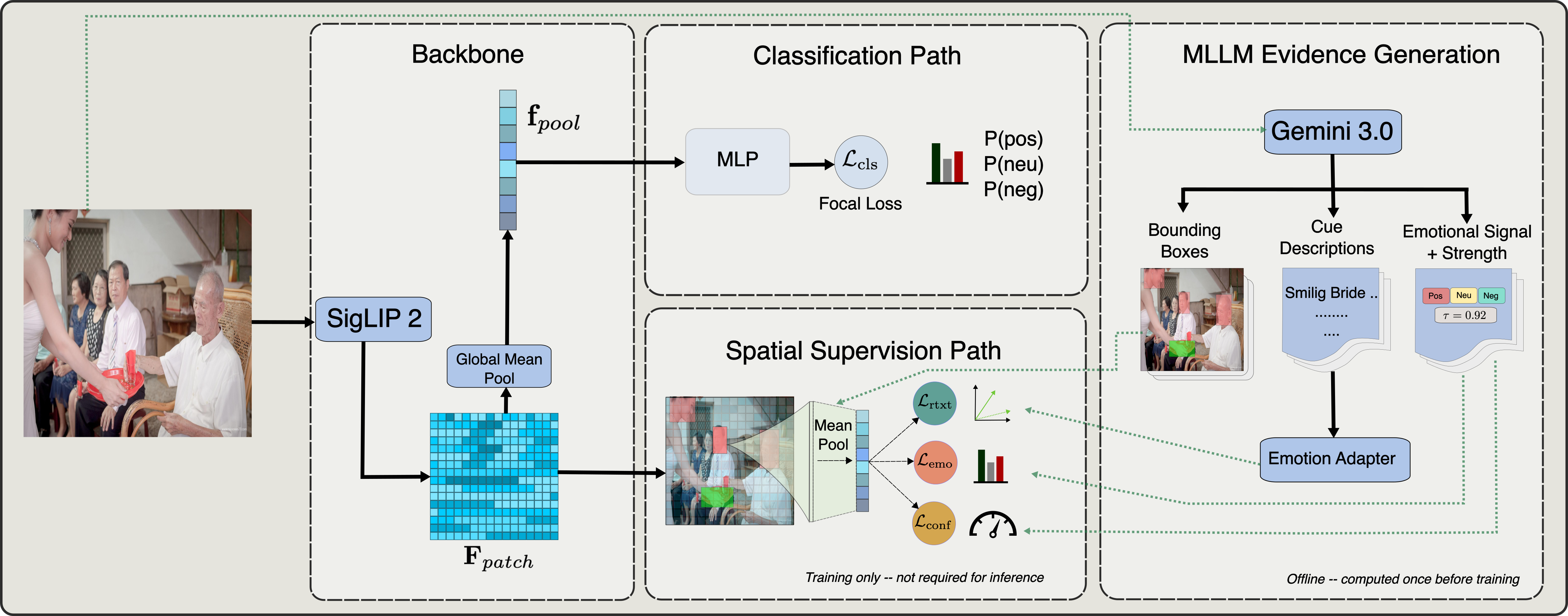}
  \caption{\small{An overview of the LG-GER framework. The VLM backbone produces patch features $\mathbf{F}_\text{patch}$ and pooled representations $\mathbf{f}_\text{pool}$. The classification path uses $\mathcal{L}_\text{cls}$ on the pooled representation. During training, patch features within evidence bounding boxes are mean-pooled and supervised by $\mathcal{L}_\text{rtxt}$, $\mathcal{L}_\text{emo}$, and $\mathcal{L}_\text{conf}$ via the spatial supervision path.}}
  \label{fig:pipeline}
\end{figure*}

\vspace{-0.1in}
\subsection{Group Emotion Recognition}
\label{sec:rw_ger}

Group emotion recognition (GER) aims to classify the collective emotional state of a group of people from a single image. The field was catalyzed by the EmotiW challenge series~\cite{dhall2015more,dhall2017individual,dhall2018emotiw}, which introduced the Group Affect Database with images labeled as positive, neutral, or negative, and the GroupEmoW dataset~\cite{guo2020graph} significantly expanded the scale and diversity of available benchmarks.

The dominant paradigm decomposes GER into detection, per-region feature extraction, and fusion. Early methods combined face-level predictions with global scene features through separate classification networks. Tan et al.~\cite{tan2017group} fused individual facial emotion CNNs with global image classifiers. Guo et al.~\cite{guo2017group} combined VGG scene features with face-level predictions. Khan et al.~\cite{khan2018group} extended this with a four-stream framework incorporating face-location attention heatmaps and scene-only features. These methods established a multi-stream, detection-dependent pipeline that persists in most subsequent work.

To better capture inter-person dynamics, several methods adopted Graph Neural Networks (GNNs) over detected entities. Guo et al.~\cite{guo2018group} extracted face, skeleton, and scene features with separate backbones and fused them via learned attention, while their later work~\cite{guo2020graph} introduced a GNN that passes messages between nodes representing detected faces, objects, and scene regions to model pairwise interactions. Wang et al.~\cite{wang2023affmech} employed an affective mechanism based on a Graph Convolutional Network (GCN) that explicitly models interpersonal emotional influence. Wang et al.~\cite{wang2022congnn} proposed a context-consistent cross-graph network that captures both inter- and intra-branch relations among face, object, and scene cues. Li et al.~\cite{li2025fehss} constructed fuzzy emotion hierarchies from detected faces and aggregated them via GCN with a Swin-L backbone.

Attention and transformer-based approaches brought more flexible feature interaction to GER. Khan et al.~\cite{khan2021caran} introduced a regional attention mechanism that estimates the importance of individual faces and objects based on their visual features and spatial context, combined with a context-aware fusion module that dynamically adjusts stream weights from image content rather than learning fixed weights. Xie et al.~\cite{xie2023dcat} proposed a dual cross-attention transformer where the most-important-person face crop and the global image exchange information through bidirectional cross-patch attention, allowing face-level and scene-level features to selectively attend to each other. Zhu et al.~\cite{zhu2023ual} took a different approach by modeling each detected individual as a Gaussian distribution and assigning uncertainty-sensitive scores to downweight occluded or ambiguous faces during aggregation. Other directions include semi-supervised training with contrastive pretraining on unlabeled data~\cite{zhang2022ssger} and hierarchical aggregation that progressively combines face-level features into group representations~\cite{fujii2020hierarchical}. Li et al.~\cite{li2026bridgeformer} proposed BridgeFormer, a prototype-subgraph transformer that constructs a heterogeneous graph over detected faces, objects, and scenes with class-level prototype anchors for long-tail amplification, achieving the highest reported accuracy on the GroupEmoW dataset~\cite{guo2020graph}.

Most recently, Zhu et al.~\cite{zhu2025scene} integrated LLM-generated scene descriptions and semantic emotion labels into a multi-stream model, which is the only prior use of language model outputs in GER and the closest prior work to ours. However, their method uses language as an additional input feature at inference time and still requires multi-stream detection.

Despite steady benchmark progress, all top-performing GER methods require explicit face or person detection at inference time, coupled with multi-stream architectures for feature fusion. Our work departs from this paradigm entirely.

\subsection{VLMs, MLLMs, and Distillation for Emotion Recognition}
\label{sec:rw_vlm_emotion}

Contrastive vision-language pretraining, such as CLIP~\cite{radford2021learning}, SigLIP~\cite{zhai2023sigmoid}, and SigLIP~2~\cite{tschannen2025siglip2}, has produced VLM backbones with strong zero-shot transfer and, in the case of SigLIP~2, patch-level spatial grounding that we exploit for our spatial supervision losses. These backbones have been rapidly adopted for emotion tasks, primarily at the individual level. CLIP-based methods adapt the vision-language space for facial expression recognition through generated textual descriptions~\cite{zhao2023dferclip}, sample-level captions~\cite{foteinopoulou2023emoclip}, LLM-derived action-unit semantics~\cite{zhao2024expclip}, or learned emotion-aware prompts~\cite{li2023cliper,zhou2024ceprompt,saadi2025peclip,chen2024finecliper}. MLLMs have also been applied directly: EmoLA~\cite{li2024emola} incorporates facial priors via instruction tuning, Emotion-LLaMA~\cite{cheng2024emotionllama} fuses audio-visual-textual inputs, and ExpLLM~\cite{lan2025expllm} uses chain-of-thought prompting for interpretable expression reasoning. All of these methods, however, target \emph{individual}-level recognition from cropped faces, single-person scenes, or per-subject video clips.

A parallel development is the use of large models as \emph{offline annotators} rather than inference-time components. Knowledge distillation~\cite{hinton2015distilling} classically transfers soft probabilities or intermediate features~\cite{romero2015fitnets} from teacher to student. More recently, MLLMs such as Gemini~\cite{gemini2025flash}, GPT-4V~\cite{achiam2023gpt4}, and LLaVA~\cite{liu2023llava} have been used to generate structured annotations at scale, offering a practical alternative to costly manual labeling. In individual emotion recognition, Exp-CLIP~\cite{zhao2024expclip} and CEPrompt~\cite{zhou2024ceprompt} distill LLM knowledge into text prompts or soft logits for cropped faces, providing semantic, class-level supervision rather than spatial.

Our work is the first to apply MLLM-guided distillation to group emotion recognition, and it does so with a different form of supervision: spatially grounded, structured evidence. Rather than transferring soft predictions or categorical descriptions, we distill bounding boxes with per-region emotion signals, evidence types, and continuous confidence scores, allowing the student to learn \emph{where} to attend, \emph{what} emotion each region conveys, and \emph{how strongly} it should influence the prediction.

%% file: 3_methodology.tex
\section{Methodology}
\label{sec:methodology}

An overview of the LG-GER framework is shown in Figure~\ref{fig:pipeline}. The framework has three phases. First, structured spatial evidence is generated offline using an MLLM (\S\ref{sec:annotation}). During training (Phase II), this evidence is distilled into a VLM backbone, employing four complementary supervision signals (\S\ref{sec:losses}). At inference (Phase III), all auxiliary heads and MLLM-derived data are discarded, leaving only the backbone with a classification head, leading to detection-free inference with no MLLM and no multi-stream fusion.

\subsection{MLLM Evidence Annotation}
\label{sec:annotation}

Current GER methods are trained with a single image-level emotion label, which lacks guidance about \emph{where} emotional cues appear or \emph{how strongly} each contributes to the overall group emotion. As shown in Figure~\ref{fig:ger_complexity}, group emotion arises from the interplay of faces, gestures, objects, and scene context, yet the training signal treats the entire image as a single, undifferentiated unit. Manually annotating region-level evidence is prohibitively expensive at scale. We use an MLLM as an offline annotator to generate spatially grounded, region-level supervision for each training image.

The MLLM is prompted to identify evidence regions in each training image and produce a structured annotation. For each image, the MLLM generates a set of evidence regions $\{e_i\}_{i=1}^{N}$. Each evidence region $e_i$ consists of five fields:
\begin{itemize}
    \item A bounding box $\mathbf{b}_i = [y_\text{min}, x_\text{min}, y_\text{max}, x_\text{max}]$ in normalized coordinates.
    \item A type label indicating the kind of evidence (e.g., faces, gestures, objects, or signs with text).
    \item A natural-language cue description $c_i$, e.g., ``smiling woman holding a trophy''.
    \item An emotion signal $s_i \in \{\text{positive}, \text{neutral}, \text{negative}, \text{ambiguous}\}$.
    \item A signal strength $\tau_i \in [0, 1]$ measuring how strongly the cue signals its emotion.
\end{itemize}

Since the MLLM produces evidence at multiple granularities, from tightly localized regions like individual faces to diffuse scene-level descriptions, we retain only localized evidence types, i.e., faces, gestures, objects, and signs, whose bounding boxes occupy a small fraction of the spatial grid. Diffuse types such as pose, interaction, and environment are excluded because their large bounding boxes yield region features close to the global average, providing little spatial discrimination. Coverage statistics are reported in \S\ref{sec:impl_details}.

\subsection{Model Architecture}
\label{sec:architecture}

We use a pretrained vision-language transformer, e.g., SigLIP~2~\cite{tschannen2025siglip2}, as the backbone, because the region-text grounding loss (\S\ref{sec:losses}) requires visual features that are already aligned with textual representations in a shared embedding space. A vision-language backbone provides this alignment natively, avoiding the need for a separate alignment step or projection layer. Most of the backbone is kept frozen to preserve pretrained representations, with only the final transformer blocks fine-tuned to adapt to the GER task.

\textbf{Dual output streams.}
The backbone produces a spatial grid of patch features $\mathbf{F}_\text{patch} \in \mathbb{R}^{S \times S \times D}$ from each input image, where $S$ is the patch grid size and $D$ is the embedding dimension. A pooled representation $\mathbf{f}_\text{pool} \in \mathbb{R}^{D}$ is obtained by mean-pooling $\mathbf{F}_\text{patch}$ over the spatial dimensions. These two outputs serve different roles in the framework: the pooled representation drives classification, while the patch features enable spatially localized supervision.

\textbf{Classification head.}
A lightweight MLP maps the pooled representation to class logits for the three emotion categories (positive, neutral, and negative). Focal loss~\cite{lin2017focal} is used with class-balanced weights to address the inherent class imbalance in GER datasets:
\begin{equation}
\label{eq:cls}
\mathcal{L}_\text{cls} = \ell_\text{focal}(\mathbf{f}_\text{pool},\; c)
\end{equation}
where $c$ is the ground-truth class. The focal loss is defined as:
\begin{equation}
\label{eq:focal}
\ell_\text{focal}(p_c) = -\alpha_c\, (1 - p_c)^{\gamma} \log p_c
\end{equation}
where $p_c$ is the predicted probability for class $c$, $\alpha_c$ is a class-balancing weight inversely proportional to class frequency, and $\gamma$ is a focusing parameter that down-weights easy examples.

\begin{figure}[t]
  \centering
  \includegraphics[width=\columnwidth]{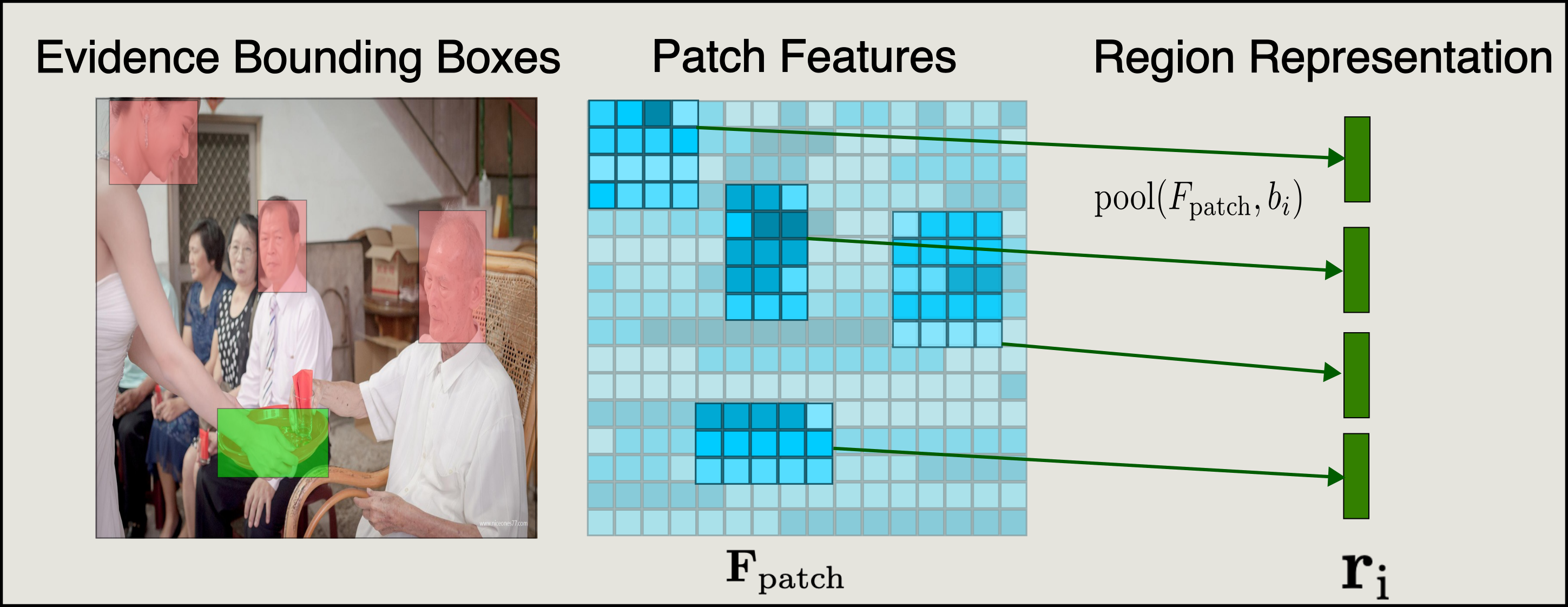}
  \caption{\small{Spatial supervision heads. Patch features $\mathbf{F}_\text{patch}$ within each evidence bounding box are mean-pooled into region features $\mathbf{r}_i$, which are supervised by three losses: region-text grounding ($\mathcal{L}_\text{rtxt}$), spatial emotion ($\mathcal{L}_\text{emo}$), and spatial confidence ($\mathcal{L}_\text{conf}$).}}
  \label{fig:spatial_losses}
\end{figure}

\textbf{Spatial supervision heads.}
Three auxiliary heads operate on the patch features during training only (Figure~\ref{fig:spatial_losses}). For each MLLM-provided evidence bounding box, we identify which patches fall within the box and mean-pool their features to obtain a region representation (\emph{region pooling}):
\begin{equation}
\label{eq:pool}
\mathbf{r}_i = \text{pool}(\mathbf{F}_\text{patch}, \mathbf{b}_i) = \frac{1}{|\mathcal{P}_i|} \sum_{j \in \mathcal{P}_i} \mathbf{f}_j
\end{equation}
where $\mathbf{f}_j \in \mathbb{R}^{D}$ is the $j$-th feature in the flattened patch grid ($j \in \{1, \ldots, S^2\}$) and $\mathcal{P}_i$ is the set of patch indices within bounding box $\mathbf{b}_i$. This routing directs spatial supervision to the relevant patch features without interfering with the pooled representation used for classification, allowing the model to develop both holistic and localized representations without gradient competition between the two objectives.

\begin{figure}[t]
  \centering
  \includegraphics[width=\columnwidth]{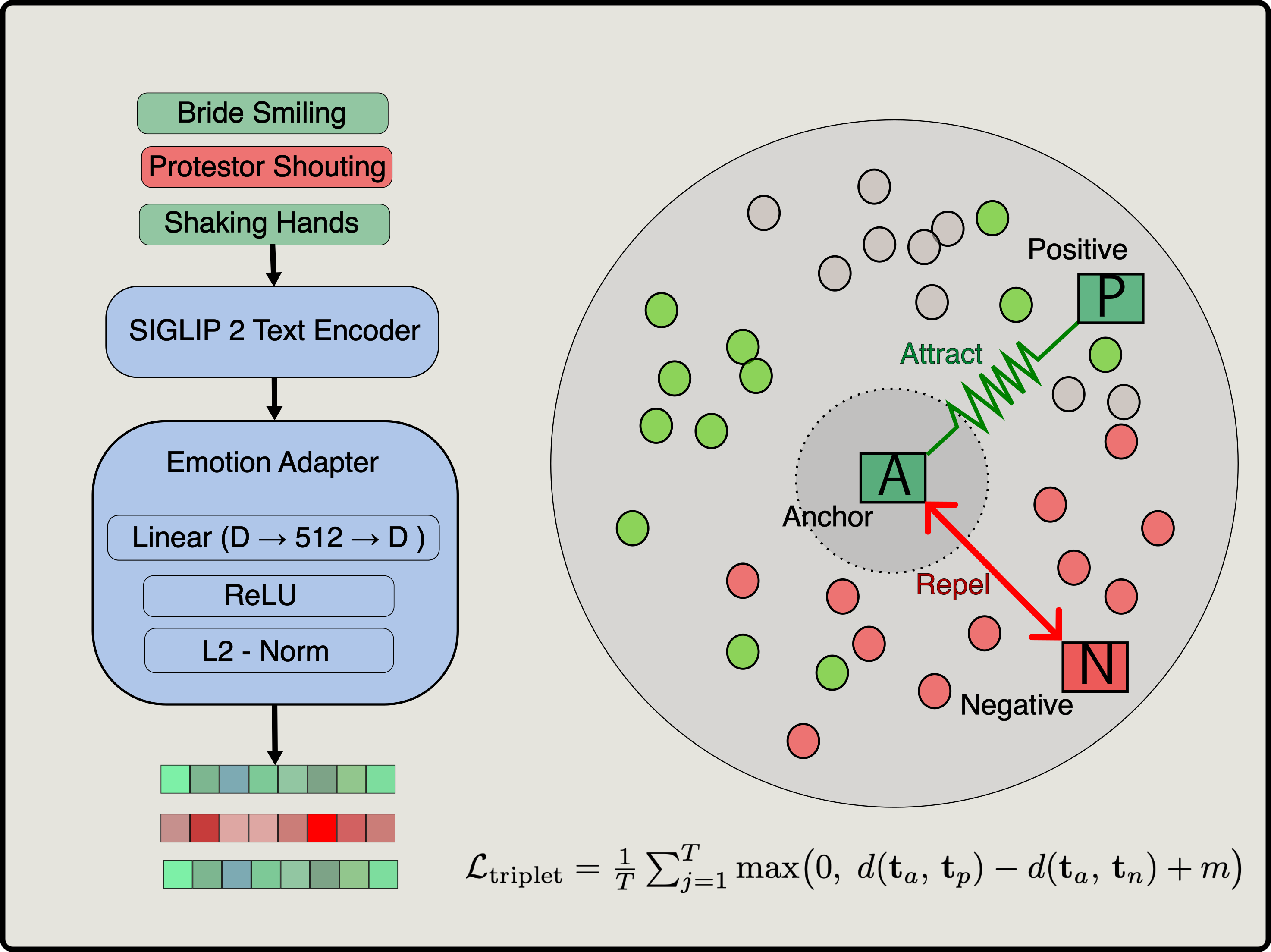}
  \caption{\small{The Emotion Adapter remaps VLM text embeddings into an emotion-discriminative space via triplet loss.}}
  \label{fig:emotion_adapter}
\end{figure}

\subsection{Emotion Adapter: Emotion-Discriminative Text Embeddings}
\label{sec:adapter}

The region-text grounding loss (\S\ref{sec:losses}) aligns visual patch features with textual embeddings of the MLLM-generated cue descriptions. This alignment is more effective when the text embeddings are separable across emotion classes. However, standard VLM text encoders are optimized for broad visual-semantic matching rather than emotional discrimination, so emotionally opposite descriptions (e.g., ``cheering joyfully'' vs.\ ``standing silently'') can end up close in embedding space when both describe human activities in similar visual contexts.

This is addressed with an \emph{Emotion Adapter} as shown in Figure~\ref{fig:emotion_adapter}: a small MLP is trained offline with a triplet loss~\cite{schroff2015facenet} to remap the VLM's text embeddings into an emotion-discriminative space. For each cue description $c_i$, the VLM text encoder produces an embedding $\mathbf{e}_i \in \mathbb{R}^{D}$, and the adapter yields an emotion-discriminative embedding $\mathbf{t}_i$:
\begin{equation}
\label{eq:adapter}
\mathbf{e}_i = \text{enc}(c_i), \quad \mathbf{t}_i = g_\phi(\mathbf{e}_i)
\end{equation}
where $g_\phi$ is the adapter MLP followed by L2 normalization and $\mathbf{t}_i \in \mathbb{R}^{D}$. Since the VLM's text encoder and vision backbone have the same embedding dimension $D$, the adapted text embeddings can be directly compared with region features through cosine similarity without any additional projection.

\textbf{Triplet loss.}
The adapter $g_\phi$ is trained with a triplet margin loss over triplets $(a, p, n)$ drawn from the MLLM-generated cue descriptions:
\begin{equation}
\label{eq:triplet}
\mathcal{L}_\text{triplet} = \frac{1}{T} \sum_{j=1}^{T} \max\!\big(0,\; d(\mathbf{t}_a,\, \mathbf{t}_p) - d(\mathbf{t}_a,\, \mathbf{t}_n) + m\big)
\end{equation}
where $d(\cdot, \cdot)$ denotes Euclidean distance; $\mathbf{t}_a$, $\mathbf{t}_p$, and $\mathbf{t}_n$ are the adapted embeddings of the anchor, positive, and negative cue descriptions, respectively; $m$ is the margin; and $T$ is the number of triplets.

\textbf{Training data.}
Each MLLM-generated cue carries a description and an emotion signal. Only cues whose emotion signal matches the image's ground-truth label are retained. Then triplets are formed, where the anchor and positive are drawn from the same emotion class and the negative is sampled uniformly at random from a different class. The adapter produces an order-of-magnitude improvement in embedding class separation as shown in Figure~\ref{fig:adapter_tsne}.

\subsection{Spatial Supervision Losses}
\label{sec:losses}

The MLLM-generated evidence provides region-level training signals that complement the image-level classification loss. Each spatial loss addresses a different aspect of the supervision: \emph{what} textual description a region corresponds to, \emph{which} emotion it signals, and \emph{how confidently} it signals that emotion. During training, three spatial heads operate on the region features $\mathbf{r}_i$ (Eq.~\ref{eq:pool}) extracted from the MLLM-provided evidence boxes. At inference, all three heads are discarded.

\subsubsection{Region-Text Grounding Loss ($\mathcal{L}_\text{rtxt}$)}

For each evidence region with cue description $c_i$, we align the corresponding patch features with the textual content so that the backbone develops spatially localized representations rather than relying solely on holistic image-level features. Specifically, the region-text grounding loss is calculated as minimizing the cosine distance between the region feature $\mathbf{r}_i$ and the adapted text embedding $\mathbf{t}_i$:
\begin{equation}
\label{eq:rtxt}
\mathcal{L}_\text{rtxt} = \frac{1}{N} \sum_{i=1}^{N} \left(1 - \cos\!\left(\mathbf{r}_i,\; \mathbf{t}_i\right)\right)
\end{equation}
where $N$ is the total number of evidence regions in the batch and $\cos(\cdot, \cdot)$ denotes cosine similarity. This loss operates on patch-level features rather than the pooled representation, allowing the backbone to develop both holistic and local representations simultaneously.

\subsubsection{Spatial Emotion Loss ($\mathcal{L}_\text{emo}$)}

The spatial emotion head classifies the emotional valence of each evidence region directly from its visual features, supervising the backbone to recognize local emotional patterns at the patch level. A lightweight MLP classifies each region's emotion signal $s_i \in \{\text{positive}, \text{neutral}, \text{negative}\}$, excluding regions labeled as ``ambiguous'':
\begin{equation}
\label{eq:emo}
\mathcal{L}_\text{emo} = \frac{1}{|\mathcal{E}|} \sum_{i \in \mathcal{E}} \ell_\text{focal}(\mathbf{r}_i,\; s_i)
\end{equation}
where $\mathcal{E}$ is the set of evidence regions with non-ambiguous emotion signals and $\ell_\text{focal}$ is the focal loss applied to the head's output.

\subsubsection{Spatial Confidence Loss ($\mathcal{L}_\text{conf}$)}

The spatial confidence head estimates the MLLM's signal strength $\tau_i \in [0, 1]$ for each region, learning to assess how strongly each region signals its emotion. A lightweight MLP regresses the signal strength from the region feature using binary cross-entropy (BCE):
\begin{equation}
\label{eq:conf}
\mathcal{L}_\text{conf} = \frac{1}{N} \sum_{i=1}^{N} \text{BCE}(\hat{\tau}_i,\; \tau_i)
\end{equation}
where $\hat{\tau}_i = \sigma(z_i)$ is the predicted signal strength and $z_i$ is the head's raw logit; $\tau_i \in [0,1]$ is the MLLM-assigned target. The BCE is defined as:
\begin{equation}
\label{eq:bce}
\text{BCE}(\hat{\tau},\; \tau) = -\tau \log \hat{\tau} - (1 - \tau) \log (1 - \hat{\tau})
\end{equation}

\subsubsection{Total Loss}

The four losses are combined with dynamically learned weights:
\begin{equation}
\label{eq:total}
\mathcal{L} = w_\text{cls}\,\mathcal{L}_\text{cls} + w_\text{rtxt}\,\mathcal{L}_\text{rtxt} + w_\text{emo}\,\mathcal{L}_\text{emo} + w_\text{conf}\,\mathcal{L}_\text{conf}
\end{equation}
where the weights $\{w_k\}$ are managed by GradNorm~\cite{chen2018gradnorm}. Dynamic balancing is necessary because the four losses operate on different feature pathways (pooled vs.\ patch) and target different signal types (classification, alignment, or regression), producing gradient magnitudes that vary by orders of magnitude. GradNorm measures per-task gradient norms on the shared backbone and adjusts the weights to maintain a target ratio, keeping the losses balanced during training.

\textbf{Per-image normalization and area weighting.}
The spatial loss formulations in Eqs.~\ref{eq:rtxt}--\ref{eq:conf} use simple averaging for clarity. In practice, images vary in the number and size of evidence regions, so without normalization, images with many annotations would dominate the batch loss. To deal with this issue, per-image normalization and area weighting are applied. Per-image normalization ensures that each image's contribution reflects evidence difficulty rather than annotation count. Within each image, each region is weighted inversely by its bounding-box area so that small, focused regions are not dominated by large ones:
\begin{equation}
\label{eq:norm}
\mathcal{L}_\text{spatial} = \frac{1}{B} \sum_{b=1}^{B} \frac{1}{N_b} \sum_{i=1}^{N_b} w_i^\text{area} \cdot \ell_i, \quad w_i^\text{area} \propto \frac{1}{\text{area}(\mathbf{b}_i)}
\end{equation}
where $B$ is the batch size, $N_b$ is the number of evidence regions in image $b$, $\ell_i$ is the per-region loss, and the area weights are normalized per-image such that $\sum_i w_i^\text{area} = N_b$.

%% file: 4_experiment.tex
\section{Experiments}
\label{sec:experiment}

\subsection{Experimental Setup}
\label{sec:setup}

\textbf{Datasets.}
The LG-GER is evaluated on two GER benchmarks.
\emph{GroupEmoW}~\cite{guo2020graph} is the largest static group emotion dataset, containing 11,127 training, 3,178 validation, and 1,589 test images.
\emph{GAF~3.0}~\cite{dhall2018emotiw}, from the EmotiW 2018 challenge, contains 9,815 training and 4,346 validation images. Since its test labels are not publicly available, all post-competition evaluation uses the validation split.
Both datasets use three classes: positive, neutral, and negative.

\textbf{Training.}
The LG-GER is optimized with AdamW (lr\,=\,$1 \times 10^{-4}$, weight decay\,=\,0.01) using cosine annealing with a 2-epoch linear warmup for 30 epochs, batch size~32, bf16 mixed precision, and gradient clip norm of 1.0.
Data augmentation includes random resized crop ($256$, scale $[0.7, 1.0]$), horizontal flip ($0.5$), color jitter ($0.3, 0.3, 0.3$), and random augment ($2, 9$).

\textbf{Evaluation.}
We report the GER accuracy on the GroupEmoW test set and the GAF~3.0 validation split.
For wide images (aspect ratio $\geq 1.7$), we average logits over multiple crops along the larger dimension; all other images use a single center crop.

\subsection{Implementation Details}
\label{sec:impl_details}

\textbf{Backbone.}
We use SigLIP~2 ViT-L/16-256~\cite{tschannen2025siglip2} as the vision backbone ($D{=}1024$, $S{=}16$). Given a $256 \times 256$ input, the model produces a pooled representation $\mathbf{f}_\text{pool} \in \mathbb{R}^{D}$ and a grid of patch features $\mathbf{F}_\text{patch} \in \mathbb{R}^{S \times S \times D}$. The last two transformer blocks are unfrozen during training with a $0.1\times$ learning rate; all earlier layers remain frozen.

\textbf{Classification head.}
A 2-layer MLP ($D \to 512 \to 3$, GELU activation, dropout 0.1) maps the pooled representation to three class logits, trained with the focal loss ($\gamma{=}2.0$, class-balanced $\alpha$).

\textbf{Spatial supervision heads.}
The spatial emotion head is an MLP ($D \to 256 \to 3$, ReLU, dropout 0.1) that classifies per-region emotion signals. The spatial confidence head is an MLP ($D \to 256 \to 1$, ReLU, dropout 0.1) that regresses signal strength via BCE.

\textbf{Emotion adapter.}
A 2-layer MLP ($D \to 512 \to D$, ReLU, L2-normalized output) is trained offline to remap SigLIP~2 text embeddings into an emotion-discriminative space.
We construct triplets from MLLM-generated cue descriptions. We first filter to retain cues whose emotion signal matches the ground-truth label (66.3\% pass rate), yielding ${\sim}$120K aligned cues, then mine 100K triplets (90K train / 10K validation) with triplet margin $m{=}0.3$ and uniform negative sampling.
Training converges in 20 epochs (AdamW, lr\,=\,$10^{-3}$) with $>$99\% of triplets satisfying the margin.

\textbf{MLLM annotation.}
We use Gemini 3.0 Flash~\cite{gemini2025flash} for evidence annotation, chosen for its native bounding-box grounding and API-level JSON schema enforcement.

\textbf{Evidence selection.}
We analyze the mean patch-grid coverage per evidence type. The diffuse types, i.e., \textit{pose} (26.7\%), \textit{interaction} (31.5\%), \textit{environment} (23.2\%), span too much of the patch grid for effective spatial supervision. Hence, we retain localized types: \textit{face} (8.3\%), \textit{gesture} (10.2\%), \textit{object} (11.6\%), and \textit{sign with text} (14.2\%). After selection, each image retains 3--8 evidence regions covering approximately 12--15\% of the patch grid.

\textbf{Loss balancing.}
GradNorm~\cite{chen2018gradnorm} manages loss weights with task multipliers $m_\text{cls}{=}2.5$ and $m_\text{conf}{=}2.0$ to prioritize classification and confidence signals. A 2-epoch curriculum warmup activates only $\mathcal{L}_\text{cls}$ and $\mathcal{L}_\text{rtxt}$ before introducing the spatial emotion and confidence losses.

\textbf{Attention extraction.}
For qualitative analysis, we extract Grad-CAM~\cite{selvaraju2017gradcam} maps from the first LayerNorm of the final transformer block as illustrated in Figure~\ref{fig:attention_evidence}.

\begin{figure}[t]
  \centering
  \includegraphics[width=\columnwidth]{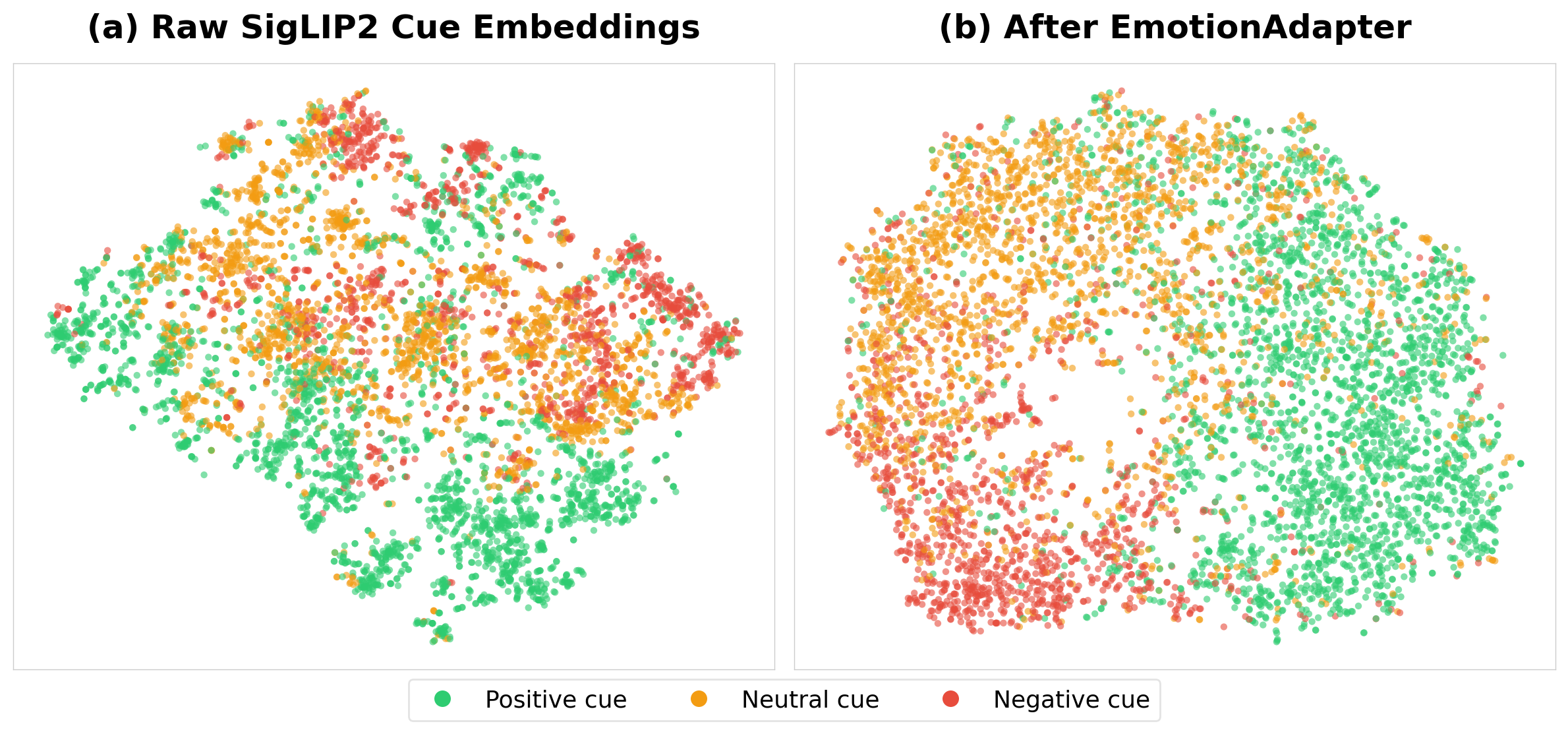}
  \caption{\small{t-SNE of cue-text embeddings. (a)~Raw SigLIP2 embeddings. (b)~After Emotion Adapter.}}
  \label{fig:adapter_tsne}
\end{figure}

\subsection{Results}
\label{sec:results}

\subsubsection{Comparison with the State-of-the-art Methods}

Table~\ref{tab:sota} compares LG-GER with recent methods on both benchmarks. All prior methods require multi-component inference pipelines with explicit detection, while LG-GER requires no detection at inference. On GAF~3.0, LG-GER achieves 84.08\% validation accuracy, the highest reported result on this dataset, surpassing BridgeFormer~\cite{li2026bridgeformer} (83.58\%) by +0.50\%. On GroupEmoW, LG-GER reaches 92.39\% test accuracy, the second-highest result after BridgeFormer (94.77\%), while with higher accuracy for negative and neutral emotions. Furthermore, LG-GER is the only method that does not require detection or multi-stream fusion at inference. 

\begin{table}[t]
\centering
\caption{\small{Comparison with the state-of-the-art methods. Specifically, test accuracy is reported on GroupEmoW; post-challenge validation accuracy is reported on GAF~3.0. All SOTA methods require detection-based multi-stream inference. ``---'' indicates the method was not evaluated on that dataset.}}
\label{tab:sota}
\small
\begin{tabular}{@{}lcc@{}}
\toprule
Method & GEW (\%) & GAF~3.0 (\%) \\
\midrule
GNN~\cite{guo2020graph}                & 89.14 & 79.08 \\
CARAN~\cite{khan2021caran}             & 90.18 & 79.13 \\
FANet~\cite{wang2023affmech}           & 90.06 & --- \\
DCAT~\cite{xie2023dcat}                & 90.47 & 79.20 \\
VSIM~\cite{zhu2025scene}               & 91.18 & 81.42 \\
FEHSS~\cite{li2025fehss}               & 91.77 & --- \\
BridgeFormer~\cite{li2026bridgeformer} & \textbf{94.77} & 83.58 \\
\midrule
\textbf{LG-GER (ours)}                & 92.39 & \textbf{84.08} \\
\bottomrule
\end{tabular}
\end{table}

\begin{figure*}[t]
  \centering
  \includegraphics[width=\textwidth]{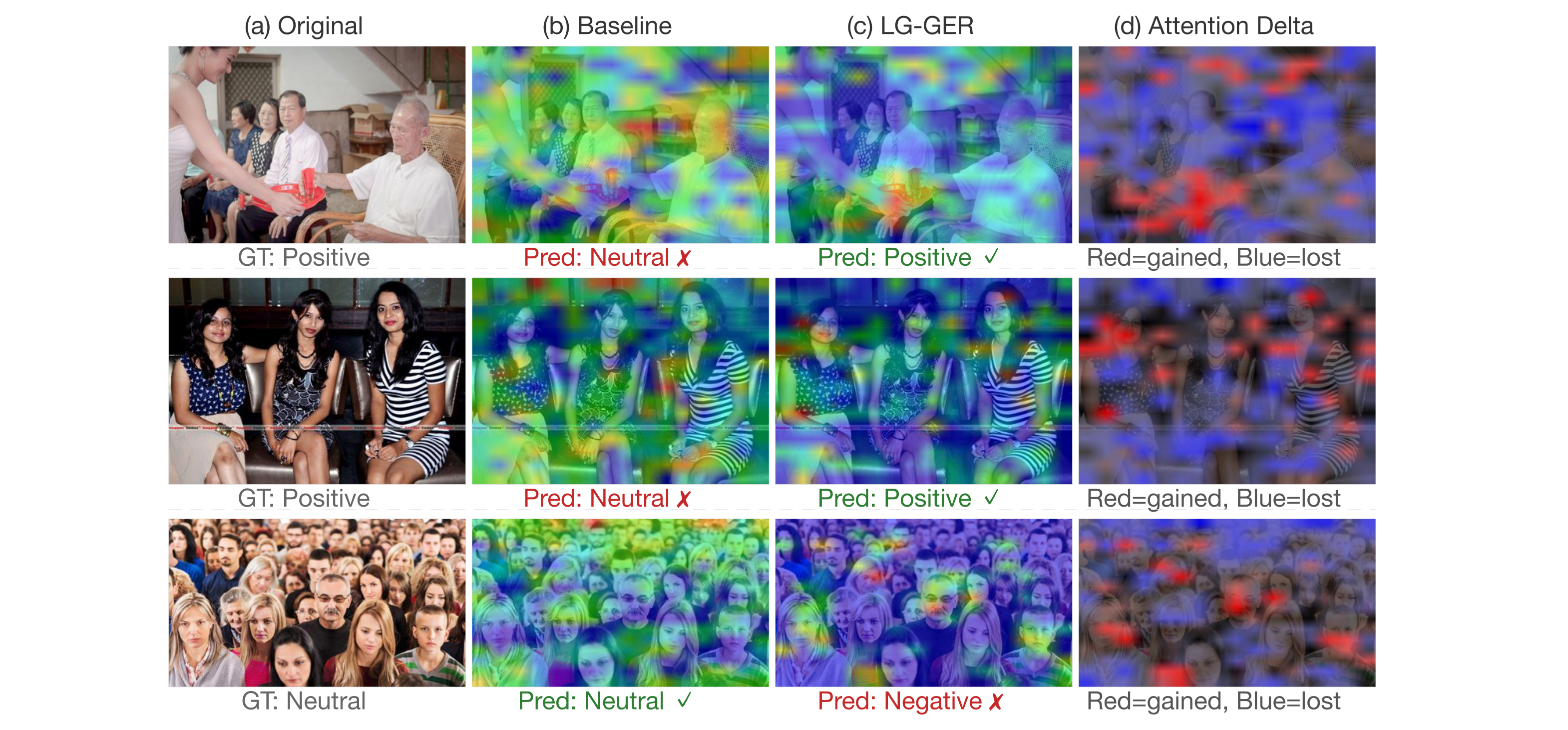}
  \caption{\small{Grad-CAM attention maps for the baseline and LG-GER on GroupEmoW test images. Red/blue in the difference maps indicate gained/lost attention. Top and middle: LG-GER correctly focuses on emotionally informative regions. Bottom: a failure case where focused attention misclassifies a neutral scene.}}
  \label{fig:attention_evidence}
\end{figure*}

\subsubsection{Ablation Study}

Table~\ref{tab:ablation} presents the component ablation on both benchmarks.

\begin{table}[t]
\centering
\caption{\small{Ablation study on SigLIP~2 ViT-L/16-256. Each row indicates which losses are active. $\dagger$\,uses raw VLM text embeddings (no Emotion Adapter); all other rows use the Emotion Adapter. Row~(f) is the full LG-GER method.}}
\label{tab:ablation}
\small
\begin{tabular}{@{}lcc@{}}
\toprule
Variant & GEW (\%) & GAF~3.0 (\%) \\
\midrule
Gemini 3.0 Flash (zero-shot)~\cite{gemini2025flash} & 84.7 & 75.0 \\
\midrule
(a) $\mathcal{L}_\text{cls}$                                                                              & 91.32 & 83.06 \\
(b) $\mathcal{L}_\text{cls} + \mathcal{L}_\text{rtxt}^{\dagger}$                                         & 91.25 & 83.16 \\
(c) $\mathcal{L}_\text{cls} + \mathcal{L}_\text{rtxt}$                                                   & 91.76 & 83.73 \\
(d) $\mathcal{L}_\text{cls} + \mathcal{L}_\text{rtxt} + \mathcal{L}_\text{emo}$                          & 91.88 & 83.89 \\
(e) $\mathcal{L}_\text{cls} + \mathcal{L}_\text{rtxt} + \mathcal{L}_\text{conf}$                         & 91.88 & 83.78 \\
\textbf{(f) LG-GER: $\mathcal{L}_\text{cls} + \mathcal{L}_\text{rtxt} + \mathcal{L}_\text{emo} + \mathcal{L}_\text{conf}$} & \textbf{92.39} & \textbf{84.08} \\
\bottomrule
\end{tabular}
\end{table}

\textbf{Baseline.}
Row~(a) benefits from the SigLIP~2 backbone and training choices including focal loss with class-balanced weights, cosine annealing, and random augmentation.
The full method (Row~f) improves over this baseline by +1.07\% on GroupEmoW and +1.02\% on GAF~3.0, respectively.

\textbf{Gemini zero-shot comparison.}
Gemini 3.0 Flash achieves 84.7\% on GroupEmoW and 75.0\% on GAF~3.0 as a zero-shot classifier.
LG-GER improves over these by +7.7\% and +9.1\%, respectively, suggesting that the distilled spatial evidence provides a stronger training signal than the teacher's direct classification.

\begin{figure}[t]
  \centering
  \includegraphics[width=\columnwidth]{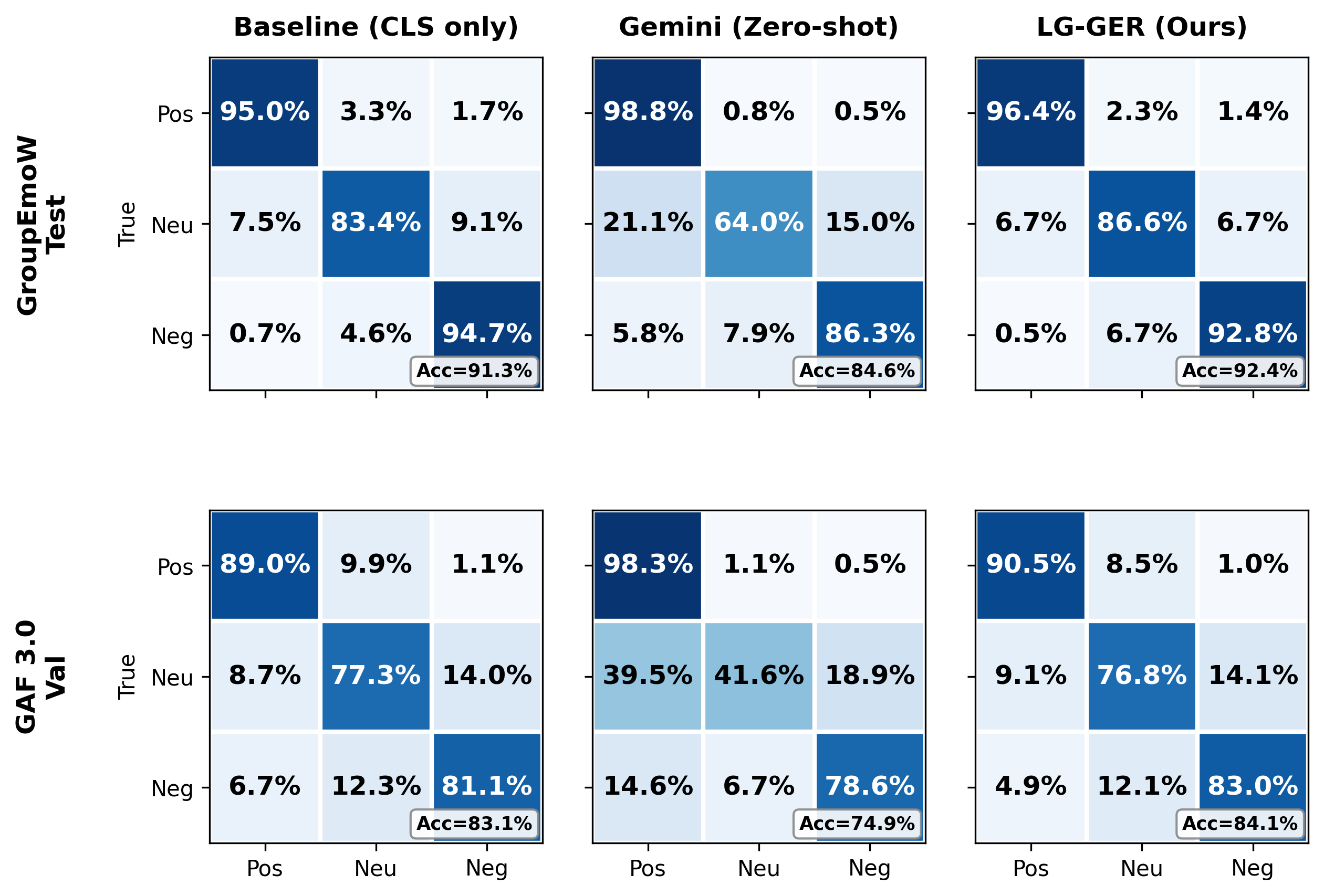}
  \caption{\small{Confusion matrices on GroupEmoW test (top) and GAF~3.0 validation (bottom) for the baseline (Row~a), Gemini zero-shot, and LG-GER (Row~f) in Table~\ref{tab:ablation}.}}
  \label{fig:confusion_matrices}
\end{figure}

\subsubsection{Emotion Adapter Effectiveness}

We measure the adapter's impact using class-separation margin, defined as the difference between average intra-class cosine similarity and average inter-class cosine similarity across the three emotion classes. Raw SigLIP~2 text embeddings exhibit a margin of 0.023, barely above the noise floor in the 1024-dimensional space. The Emotion Adapter increases this margin to 0.405, an 18$\times$ improvement as shown in Figure~\ref{fig:adapter_tsne}. Inter-class cosine similarity drops from 0.556 to 0.063, with the positive-negative pair reaching $-$0.092 after adaptation, indicating opposite directions on the unit hypersphere. This is consistent with the ablation study in Table~\ref{tab:ablation}: Row~(b), which uses raw embeddings, hurts GroupEmoW accuracy ($-$0.07\%), while Row~(c) with the adapter recovers the deficit and improves over the baseline by +0.44\%.

\subsubsection{Cross-Dataset Validation}

The preceding results demonstrate strong in-domain performance on both benchmarks. A natural follow-up is whether the language-guided supervision captures dataset-agnostic emotion patterns or merely fits each dataset's annotation conventions. To evaluate this, we evaluate LG-GER across datasets, i.e., training on one benchmark, while evaluating on the other without any fine-tuning. Table~\ref{tab:cross} reports accuracy under center-crop inference.

\begin{table}[t]
\centering
\caption{\small{Cross-dataset evaluation. LG-GER is trained on one benchmark and evaluated on the other without fine-tuning. GEW\,=\,GroupEmoW, GAF\,=\,GAF~3.0. All entries are accuracy~(\%).}}
\label{tab:cross}
\small
\begin{tabular}{@{}lcc@{}}
\toprule
Direction & Acc (\%) & Gap \\
\midrule
GEW $\to$ GEW (in-domain) & 92.39 & --- \\
GAF $\to$ GEW            & 88.23 & $-$4.16 \\
\midrule
GAF $\to$ GAF (in-domain) & 84.08 & --- \\
GEW $\to$ GAF              & 79.54 & $-$4.54 \\
\bottomrule
\end{tabular}
\end{table}

Both directions retain strong accuracy despite differences in annotation guidelines and image distributions.
The GAF$\to$GEW direction transfers at 88.23\%, a modest $-$4.16\% gap from in-domain, while GEW$\to$GAF reaches 79.54\% ($-$4.54\% gap).
These results suggest that the MLLM-generated spatial evidence captures emotion patterns that generalize across dataset boundaries, rather than overfitting to dataset-specific annotation conventions.

\subsection{Analysis}
\label{sec:analysis}

\textbf{Component contributions.}
The ablation in Table~\ref{tab:ablation} reveals a clear component hierarchy.
Raw SigLIP~2 text embeddings \emph{hurt} GroupEmoW performance (Row~b, $-$0.07\%), confirming that SigLIP~2's text encoder is not emotion-discriminative (\S\ref{sec:adapter}).
The Emotion Adapter recovers this deficit entirely (Row~c), consistent with the 18$\times$ improvement in embedding class separation shown in Figure~\ref{fig:adapter_tsne}.
On GroupEmoW, the full method (Row~f) achieves the best accuracy, confirming that $\mathcal{L}_\text{conf}$ and $\mathcal{L}_\text{emo}$ are complementary when sufficient training data is available.
On GAF~3.0, Row~(f) again leads, though the margin over Row~(c) is narrower (+0.35\% vs.\ +0.63\% on GroupEmoW), consistent with the smaller training set.

\textbf{Per-class analysis.}
Figure~\ref{fig:confusion_matrices} shows confusion matrices for the baseline, LG-GER, and the Gemini teacher on both datasets, respectively.
Gemini, despite near-perfect Positive accuracy, struggles with Neutral (64.0\% on GroupEmoW and 41.6\% on GAF~3.0). On GroupEmoW, LG-GER improves Positive and Neutral per-class accuracy over the baseline (+1.4\% and +3.2\% respectively), with Neutral seeing the largest gain. Negative per-class accuracy decreases slightly ($-$1.9\%) as the model trades some Negative predictions for improved Neutral discrimination. On GAF~3.0, Positive and Negative improve while Neutral decreases slightly. Neutral remains the bottleneck class across all configurations.

The confusion matrices in Figure~\ref{fig:confusion_matrices} show \emph{where} LG-GER's classification improvements concentrate; the attention maps in Figure~\ref{fig:attention_evidence} reveal \emph{why}---which spatial regions the model learns to prioritize after spatial training.

\textbf{Qualitative analysis.}
Figure~\ref{fig:attention_evidence} visualizes Grad-CAM attention maps for the baseline and LG-GER on three GroupEmoW test images. Across all three examples, a consistent pattern emerges: the baseline distributes attention broadly across the image, while LG-GER concentrates on emotionally informative regions.

In the top row, the baseline spreads attention across people, walls, and furniture, predicting Neutral.
LG-GER concentrates on the central serving interaction in a celebratory setting and correctly predicts Positive.
The difference map shows attention shifting \emph{toward} the contextual interaction (red) and \emph{away from} uninformative periphery (blue), illustrating how spatial supervision redirects the model from diffuse scanning to focused context reading.

The middle row shows a complementary pattern: selective face attention.
The baseline distributes attention across clothing and background, registering the presence of people but missing the emotional tone.
LG-GER amplifies attention on the most expressive face while suppressing background, correctly predicting Positive---prioritizing the strongest emotional signal rather than attending to all faces equally.

The bottom row presents an instructive failure.
In a crowd with mixed expressions, the baseline's broad attention integrates the overall scene tone and correctly predicts Neutral.
LG-GER concentrates on a centrally located person whose expression is ambiguously negative, tipping the prediction to Negative.
This failure highlights a tension between focused and diffuse strategies: for genuinely neutral scenes with mixed expressions, broad attention can outperform selective attention.

%% file: 5_conclusion.tex
\section{Conclusion}
\label{sec:conclusion}

We presented LG-GER, a framework that distills MLLM-generated spatial evidence into a single-stream VLM backbone for group emotion recognition.
By supervising patch features with region-text alignment, spatial emotion classification, and spatial confidence regression during training while discarding all auxiliary components at inference, LG-GER achieves the highest reported accuracy on GAF~3.0 and competitive results on GroupEmoW with detection-free inference.

Our analysis shows that spatial supervision changes model behavior: attention shifts from diffuse scanning to selective, evidence-driven focus. The cross-dataset validation also shows that the MLLM-generated spatial evidence is capable of capturing generalized emotional patterns. Future work includes stronger annotation-quality auditing and extension to video-based GER.